%% file: template.tex
\documentclass{Interspeech}

\interspeechcameraready

\usepackage{amsmath,graphicx}
\usepackage{hyperref}
\usepackage{booktabs} %
\usepackage{subcaption}
\usepackage{xcolor, colortbl}
\usepackage{xspace}
\usepackage{multirow}
\usepackage{xurl}
\usepackage{amsfonts}
\usepackage{pifont}

\newcommand{\cmark}{\ding{51}}
\newcommand{\xmark}{\ding{55}}

\usepackage{arydshln}
\usepackage{threeparttable} %


\usepackage[
    backend=biber,
    style=ieee,
    citestyle=numeric-comp,
    maxbibnames=5,
    minbibnames=4,
    maxcitenames=3,
    doi=false,isbn=false,url=false,eprint=false
]{biblatex}

\defbibheading{bibliography}[\refname]{}

\title{OWSM v4: Improving Open Whisper-Style Speech Models via Data Scaling and Cleaning}

\author[affiliation={1}]{Yifan}{Peng}
\author[affiliation={2}]{Muhammad}{Shakeel}
\author[affiliation={2}]{Yui}{Sudo}
\author[affiliation={1}]{William}{Chen}
\author[affiliation={1}]{Jinchuan}{Tian}
\author[affiliation={1}]{Chyi-Jiunn}{Lin}
\author[affiliation={1}]{Shinji}{Watanabe}

\affiliation{}{Carnegie Mellon University}{United States}
\affiliation{}{Honda Research Institute Japan}{Japan}
\email{pengyf21@gmail.com, shinjiw@ieee.org}

\usepackage{comment}

\begin{document}

\maketitle

\begin{abstract}
    The Open Whisper-style Speech Models (OWSM) project has developed a series of fully open speech foundation models using academic-scale resources, but their training data remains insufficient. This work enhances OWSM by integrating YODAS, a large-scale web-crawled dataset with a Creative Commons license. However, incorporating YODAS is nontrivial due to its wild nature, which introduces challenges such as incorrect language labels and audio-text misalignments. To address this, we develop a scalable data-cleaning pipeline using public toolkits, yielding a dataset with 166,000 hours of speech across 75 languages. Our new series of OWSM v4 models, trained on this curated dataset alongside existing OWSM data, significantly outperform previous versions on multilingual benchmarks. Our models even match or surpass frontier industrial models like Whisper and MMS in multiple scenarios. We will publicly release the cleaned YODAS data, pre-trained models, and all associated scripts via the ESPnet toolkit.\footnote{\url{https://www.wavlab.org/activities/2024/owsm/}}
\end{abstract}

\section{Introduction}

Speech foundation models (SFMs), typically trained on large amounts of data, have demonstrated state-of-the-art (SOTA) performance in various speech processing tasks~\cite{whisper, google-usm, meta-mms, nv-canary}.
A notable example is OpenAI's Whisper~\cite{whisper}, which is trained on 680 thousand to 5 million hours of audio data and supports multilingual automatic speech recognition (ASR), any-to-English speech translation (ST), spoken language identification (LID), and voice activity detection (VAD). However, Whisper does not publicly release its training data, code, and logs, leading to concerns about privacy, transparency, and reproducibility.
To advance open research, researchers from academic institutions have developed a series of fully open Whisper-style speech models (OWSM)~\cite{owsm-asru23} using publicly available data and an open source toolkit, ESPnet~\cite{espnet}.
Initial OWSM v1, v2, and v3~\cite{owsm-asru23} establish a reproducible pipeline for Whisper-style training, but their performance is limited.

\begingroup
\setlength{\tabcolsep}{6pt}
\begin{table}[t]
  \caption{ASR error rates (\%) on FLEURS. Our OWSM-CTC v4 outperforms v3.1 across all 102 languages and surpasses v3.2 in 100 languages. Here we only show languages where OWSM-CTC v4 achieves error rates below 20\%. \textbf{Bold}: The best result in each row. \colorbox{blue!15}{Blue}: Our v4 model surpasses previous OWSM.}
  \label{tab:fleurs}
  \vskip -0.1in
  \centering
  \resizebox {0.8\linewidth} {!} {
  \begin{tabular}{cccccc}
    \toprule
    \multirow{2}{*}{\textbf{Lang.}} & \multirow{2}{*}{\textbf{Metric} $\downarrow$} & \textbf{MMS} & \multicolumn{3}{c}{\textbf{OWSM-CTC}} \\
    \cmidrule(lr){4-6}
    & & \textbf{1B-all} & \textbf{v3.1} & \textbf{v3.2} & \textbf{v4 (ours)} \\
    \midrule
    \texttt{spa} & WER & 6.60 & 11.30 & 9.58 & \cellcolor{blue!15}{\textbf{5.44}} \\
    \texttt{ita} & WER & \textbf{5.85} & 13.38 & 11.25 & \cellcolor{blue!15}{5.91} \\
    \texttt{eng} & WER & 12.26 & 8.24 & 7.06 & \cellcolor{blue!15}{\textbf{6.37}} \\
    \texttt{jpn} & CER & 20.92 & 7.56 & 6.51 & \cellcolor{blue!15}{\textbf{6.43}} \\
    \texttt{kor} & CER & 18.30 & 20.09 & 17.72 & \cellcolor{blue!15}{\textbf{6.74}} \\
    \texttt{por} & WER & 8.97 & 19.66 & 16.22 & \cellcolor{blue!15}{\textbf{7.38}} \\
    \texttt{cat} & WER & 10.75 & 9.37 & 8.16 & \cellcolor{blue!15}{\textbf{7.70}} \\
    \texttt{deu} & WER & 10.45 & 14.97 & 13.17 & \cellcolor{blue!15}{\textbf{8.36}} \\
    \texttt{fra} & WER & 12.53 & 17.13 & 14.79 & \cellcolor{blue!15}{\textbf{9.67}} \\
    \texttt{ind} & WER & 13.19 & 39.76 & 33.56 & \cellcolor{blue!15}{\textbf{10.24}} \\
    \texttt{zho} & CER & 26.46 & 13.47 & 12.25 & \cellcolor{blue!15}{\textbf{10.95}} \\
    \texttt{rus} & WER & 19.79 & 18.18 & 15.68 & \cellcolor{blue!15}{\textbf{10.96}} \\
    \texttt{tha} & CER & \textbf{10.69} & 28.29 & 24.06 & \cellcolor{blue!15}{12.35} \\
    \texttt{vie} & WER & 29.96 & 71.36 & 65.15 & \cellcolor{blue!15}{\textbf{13.34}} \\
    \texttt{nld} & WER & \textbf{12.35} & 30.46 & 25.10 & \cellcolor{blue!15}{14.73} \\
    \texttt{bel} & WER & \textbf{14.84} & 18.99 & 16.38 & \cellcolor{blue!15}{15.09} \\
    \texttt{tur} & WER & 19.55 & 57.43 & 48.56 & \cellcolor{blue!15}{\textbf{15.79}} \\
    \texttt{ben} & WER & \textbf{13.19} & 18.63 & 16.33 & \cellcolor{blue!15}{15.80} \\
    \texttt{hin} & WER & \textbf{10.82} & 35.63 & 30.91 & \cellcolor{blue!15}{16.40} \\
    \texttt{glg} & WER & \textbf{10.59} & 30.06 & 24.42 & \cellcolor{blue!15}{17.41} \\
    \texttt{ukr} & WER & \textbf{18.09} & 47.53 & 40.97 & \cellcolor{blue!15}{18.39} \\
    \bottomrule
  \end{tabular}
  }
  \vskip -0.2in
\end{table}
\endgroup

\begin{figure*}[t]
    \centering
    \includegraphics[width=0.9\linewidth]{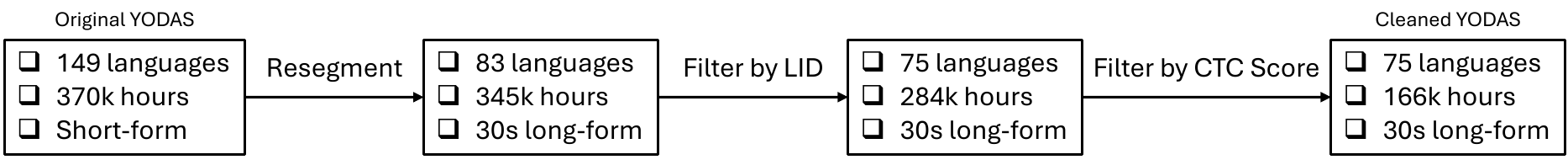}
    \vskip -0.14in
    \caption{Our data-cleaning pipeline consists of three steps: (1) realign audio and text using a pre-trained OWSM-CTC model, (2) filter data based on LID results, and (3) filter data based on CTC confidence scores.}
    \vskip -0.15in
    \label{fig:cleaning-pipeline}
\end{figure*}

Recent studies have enhanced the effectiveness and efficiency of SFMs.
One approach is improving model architectures. 
Conformer~\cite{conformer}, Branchformer~\cite{branchformer}, and Zipformer~\cite{zipformer} consistently outperform Transformer~\cite{transformer} for speech modeling.
Squeezeformer~\cite{squeezeformer}, FastConformer~\cite{fastconformer}, and SummaryMixing~\cite{summarymixing} significantly reduce the training and inference cost.
OWSM v3.1~\cite{owsm31} adopts E-Branchformer~\cite{ebf, ebf-vs-conformer} and achieves significant improvements over OWSM v3.
OWSM-CTC~\cite{owsmctc} proposes a novel non-autoregressive architecture based on hierarchical self-conditioned Connectionist Temporal Classification (CTC)~\cite{ctc}, unifying ASR, ST, and LID in a shared encoder-only model. Compared to attention-based encoder-decoder (AED) models, OWSM-CTC improves the inference speed and reduces hallucinations.

Another line of research improves training data.
Unsupervised data selection is proposed to enhance ASR systems~\cite{DBLP:conf/interspeech/LuWZHCH22, park2022unsupervised}. Data cleaning techniques are widely used when creating ASR datasets~\cite{spgispeech, gigaspeech, peoplespeech, yodas}. Inspired by this, \citeauthor{owsm32-jinchuan} filter OWSM v3.1 training data based on ASR error rates and restore punctuation and capitalization using large language models. Compared to OWSM v3.1, the resultant model, OWSM v3.2~\cite{owsm32-jinchuan}, achieves comparable ASR results and slightly better ST results, despite being trained on 15\% less data.
However, \citeauthor{owsm32-jinchuan} only consider the original v3.1 data that generally have good quality but do not include new data from other public sources. Hence, the performance gain of data filtering is marginal and inconsistent.

Inspired by the findings that scaling training data improves multilingual ASR systems~\cite{speechstew, whisper, google-usm}, we propose to enhance OWSM by integrating high-quality data from YODAS~\cite{yodas} using academic-scale resources.
YODAS is distinctive from other popular datasets such as MSR-86K~\cite{msr-86k}, LibriHeavy~\cite{libriheavy}, GigaSpeech~\cite{gigaspeech, gigaspeech2}, and MOSEL~\cite{mosel} in the following aspects:
(1) YODAS publicly releases audio files in a Creative Commons license instead of links to original sources, simplifying data downloading and providing static sources for redistribution.
(2) YODAS establishes a scalable pipeline to crawl data from the web. The current version already includes 370k hours of audio in 149 languages. Future versions can further grow.
(3) YODAS covers diverse speaking styles and acoustic environments. It also releases unsegmented long-form audio recordings, which are suitable for Whisper-style training.
However, simply adding more data without careful curation can degrade performance due to noisy annotations in the raw data. Hence, data cleaning is crucial for ensuring good quality.

Our contributions are summarized below.
\begin{itemize}
    \item We propose a scalable data-cleaning pipeline using public LID and ASR models. By applying it to YODAS, we create an ASR dataset with 166k hours of audio in 75 languages.
    \item We develop a new series of OWSM v4 models using academic-scale resources, comprising three AED models of varying sizes and one CTC model, trained on the cleaned YODAS dataset in conjunction with previous OWSM data (320k hours in total). The new models consistently and significantly outperform previous OWSM versions in multilingual ASR and LID (see \autoref{tab:fleurs} for example). Furthermore, they achieve competitive results compared to SOTA industrial models on multiple benchmarks.
    \item To advance academic research, we will publicly release our data-cleaning pipeline, the cleaned YODAS data, training code, pre-trained model weights, and training logs.
\end{itemize}

\begingroup
\setlength{\tabcolsep}{2pt}
\begin{table}[t]
  \caption{ASR WERs (\%) of OWSM v3.1 small fine-tuned on the cleaned YODAS filtered at varying thresholds $\theta_{\text{CTC}}$. LF: Long-form web presentations.}
  \label{tab:threshold-cv}
  \vskip -0.13in
  \centering
  \resizebox {\linewidth} {!} {
  \begin{tabular}{ccccccccccccc}
    \toprule
    \multirow{2}{*}{$\theta_{\text{CTC}}$} & LF & \multicolumn{11}{c}{Common Voice} \\
    & \texttt{eng} & \texttt{deu} & \texttt{eng} & \texttt{fra} & \texttt{ind} & \texttt{ita} & \texttt{nld} & \texttt{por} & \texttt{rus} & \texttt{spa} & \texttt{tur} & \texttt{vie} \\
    \midrule
    0.00 & 5.0 & 85.7 & 100+ & 100+ & 100+ & 100+ & 100+ & 100+ & 65.5 & 100+ & 100+ & 100+\\
    \rowcolor{green!10} 0.10 & \textbf{4.3} & 18.7 & 24.3 & 24.5 & 36.4 & \textbf{18.7} & 22.9 & 37.9 & \textbf{18.5} & \textbf{15.7} & \textbf{50.2} & 54.6\\
    0.15 & 4.4 & 18.2 & \textbf{24.1} & 24.1 & 35.7 & 18.8 & \textbf{22.2} & 37.9 & 19.6 & 15.8 & 52.8 & 59.9\\
    0.20 & 4.4 & 18.0 & 24.2 & 24.0 & 34.4 & 21.2 & 22.4 & 38.0 & 20.6 & 18.4 & 52.8 & 47.3\\
    0.30 & 4.6 & \textbf{17.4} & 25.0 & \textbf{22.9} & \textbf{31.9} & 20.1 & 24.8 & \textbf{37.1} & 22.4 & 17.6 & 51.1 & \textbf{47.0}\\
    \bottomrule
  \end{tabular}
  }
  \vskip -0.17in
\end{table}
\endgroup

\vskip -0.15in
\section{Proposed Method}
\vskip -0.03in

\subsection{YODAS data cleaning}
\vskip -0.05in

The raw YODAS data has not undergone a rigorous cleaning process and may contain annotation errors~\cite{yodas}. Common issues include mismatched language labels and misalignment between audio and text. Therefore, data cleaning is essential to ensure accuracy and reliability.
\autoref{fig:cleaning-pipeline} illustrates our data-cleaning pipeline consisting of the following three steps. Our scripts will be publicly released, including more implementation details.

\subsubsection{Resegmentation}
\label{subsubsec:resegmentation}
\vskip -0.05in

YODAS provides unsegmented long-form recordings, each of them is accompanied by a list of text transcriptions annotated with start and end timestamps. 
However, these timestamps can be inaccurate. 
Consequently, our first step is to realign the audio and text using the CTC segmentation algorithm~\cite{ctcsegmentation}. For this purpose, we employ the publicly available OWSM-CTC v3.2 model\footnote{\href{https://huggingface.co/espnet/owsm_ctc_v3.2_ft_1B}{https://huggingface.co/espnet/owsm\_ctc\_v3.2\_ft\_1B}}, which supports only a subset of the languages present in YODAS.
Following realignment, the long-form audio recordings are segmented into shorter utterances, each with a maximum duration of 30 seconds. Utterances that consist exclusively of non-speech elements, such as music, are removed. The processed dataset comprises 345k hours of audio across 83 languages.
Additionally, after CTC segmentation, each short utterance is assigned a confidence score, which quantifies the alignment quality between the audio and the corresponding text. This confidence score is subsequently utilized to filter low-quality data, as discussed in Section~\ref{subsubsec:ctc-score-filtering}.

\begingroup
\setlength{\tabcolsep}{3pt}
\begin{table}[t]
  \caption{Durations (in k hours) for the top 10 languages in the cleaned YODAS filtered at varying thresholds $\theta_{\text{CTC}}$.}
  \label{tab:yodas-stats}
  \vskip -0.13in
  \centering
  \resizebox {\linewidth} {!} {
  \begin{tabular}{crrrrrrrrrrr}
    \toprule
    $\theta_{\text{CTC}}$ & Total & \texttt{eng} & \texttt{spa} & \texttt{rus} & \texttt{por} & \texttt{kor} & \texttt{fra} & \texttt{deu} & \texttt{ita} & \texttt{vie} & \texttt{ind} \\
    \midrule
    0.00 & 283.6 & 129.0 & 30.7 & 25.8 & 18.2 & 17.5 & 15.5 & 12.2 & 9.3 & 7.8 & 5.2 \\
    \rowcolor{green!10} 0.10 & 166.4 & 74.6 & 17.3 & 15.7 & 10.8 & 10.9 & 8.6 & 7.1 & 5.6 & 4.6 & 3.3\\
    0.15 & 118.5 & 51.5 & 12.2 & 11.1 & 8.1 & 8.3 & 6.2 & 5.0 & 4.1 & 3.5 & 2.5\\
    0.20 & 84.8 & 35.7 & 8.6 & 7.8 & 6.1 & 6.4 & 4.5 & 3.6 & 3.0 & 2.6 & 1.9\\
    0.30 & 43.0 & 17.0 & 4.2 & 3.7 & 3.4 & 3.7 & 2.3 & 2.0 & 1.5 & 1.5 & 1.1\\
    \bottomrule
  \end{tabular}
  }
  \vskip -0.15in
\end{table}
\endgroup

\subsubsection{LID-based filtering}
\label{subsubsec:lid-filtering}
\vskip -0.05in

We observe that certain utterances have incorrect language labels. To address this issue, we perform LID on both the audio and text using public models.
Specifically, the text-based LID model\footnote{\href{https://fasttext.cc/docs/en/language-identification.html}{https://fasttext.cc/docs/en/language-identification.html}} is sourced from fastText~\cite{joulin2016bag, joulin2016fasttext}, while the spoken LID model is based on ECAPA-TDNN\footnote{\href{https://huggingface.co/speechbrain/lang-id-voxlingua107-ecapa}{https://huggingface.co/speechbrain/lang-id-voxlingua107-ecapa}}, developed by SpeechBrain~\cite{speechbrain}.
We retain only those utterances for which the original language label matches both the predicted language from the text and the predicted language from the audio. Applying this filtering step results in a dataset comprising 284k hours of audio across 75 languages, as shown in \autoref{fig:cleaning-pipeline}.

\subsubsection{CTC-score-based filtering}
\label{subsubsec:ctc-score-filtering}
\vskip -0.05in

The final step removes utterances with low-quality audio-text alignments, as indicated by the CTC score calculated in Section~\ref{subsubsec:resegmentation}.
The CTC confidence score is language-dependent; therefore, we rank the scores of short utterances within each language and select a relative threshold (quantile) $\theta_{\text{CTC}}$. 
For each long-form utterance, if any of its constituent short utterances fall within the lowest $\theta_{\text{CTC}}$ quantile, the entire utterance will be discarded.
Different threshold values yield varying amounts of retained data.
To identify a suitable threshold, we fine-tune a pre-trained small-sized OWSM v3.1 (367M)~\cite{owsm31} on the cleaned YODAS data filtered at different thresholds. We then evaluate them on Common Voice~\cite{commonvoice} for short-form ASR and a web presentation corpus for long-form ASR, as shown in \autoref{tab:threshold-cv}.

When $\theta_{\text{CTC}} = 0.00$, no filtering is applied, and all 284k hours of audio after LID filtering are used for fine-tuning. However, the performance on Common Voice is poor and unstable. The decoding often gets stuck in repetitions of a few tokens, leading to word error rates (WER) exceeding 100\%. This observation confirms the presence of substantial misalignment issues within the raw YODAS data.

Conversely, applying CTC-score-based filtering ($\theta_{\text{CTC}} > 0$) yields significant improvements, demonstrating the effectiveness of data cleaning.
Performance trends vary across different test sets.  In some cases, increased data removal leads to better performance, while in others, the opposite trend is observed.
Although finer-grained filtering could potentially optimize performance for individual languages, we opt for a threshold of $\theta_{\text{CTC}} = 0.10$. This value retains the majority of the data while providing generally good performance across languages.
This filtering process results in 166k hours of audio spanning 75 languages, as illustrated in the final panel of \autoref{fig:cleaning-pipeline}. The durations of the top 10 languages are presented in \autoref{tab:yodas-stats}. Similar to the raw YODAS data, the distribution across languages is highly imbalanced. English constitutes the largest share, whereas many other languages continue to be underrepresented.
For simplicity, in this work, we keep the original distribution without any resampling.

\begingroup
\setlength{\tabcolsep}{2pt}
\begin{table}[t]
  \caption{Configurations. Models are categorized into three types based on their level of openness, including the availability of pre-trained weights, data details, and training code \& logs.}
  \label{tab:configs}
  \vskip -0.13in
  \centering
  \resizebox {\linewidth} {!} {
  \begin{threeparttable}
  \begin{tabular}{lcccccccc}
    \toprule
    \multirow{2}{*}{\textbf{Model Name}} & \multicolumn{3}{c}{\textbf{Openness}} & \textbf{Model} & \multicolumn{2}{c}{\textbf{Data (h)}} & \textbf{GPU} & \textbf{\# of} \\
    \cmidrule{2-4}
    \cmidrule{6-7}
& {\footnotesize \textbf{Weights}} & {\footnotesize \textbf{Data}} & {\footnotesize \textbf{Logs}} & \textbf{Size} & \textbf{ASR} & \textbf{ST} & \textbf{Hrs} & \textbf{Lang.} \\
    \midrule
    \rowcolor{gray!30} \multicolumn{9}{c}{\textbf{Open-weight models}}\\
    \rowcolor{gray!10} Whisper base~\cite{whisper} & \cmark & \xmark & \xmark & 0.07B & 555k  & 125k  & unk. & 99\\
    \rowcolor{gray!10} Whisper small~\cite{whisper} & \cmark & \xmark & \xmark & 0.24B & 555k  & 125k  & unk. & 99\\
    \rowcolor{gray!10} Whisper medium~\cite{whisper} & \cmark & \xmark & \xmark & 0.77B & 555k  & 125k  & unk. & 99\\
    \rowcolor{gray!10} Whisper large v3~\cite{whisper} & \cmark & \xmark & \xmark & 1.55B & \multicolumn{2}{c}{5M } & unk. & 100\\
    \rowcolor{gray!10} Parakeet-CTC~\cite{fastconformer} & \cmark & \xmark & \xmark & 1.06B & 64k & - & unk. & 1 \\
    \rowcolor{gray!10} Canary~\cite{nv-canary} & \cmark & \xmark & \xmark & 1.02B & 82k & 66k & 6.1k\tnote{*} & 4\\
    \rowcolor{brown!40} \multicolumn{9}{c}{\textbf{Open-weight, open-data models}}\\
    \rowcolor{brown!20} MMS-fl102~\cite{meta-mms} & \cmark & \cmark\tnote{\S} & \xmark & 0.97B & 1.4k\tnote{\textparagraph} & - & unk. & 102\\
    \rowcolor{brown!20} MMS-all~\cite{meta-mms} & \cmark & \cmark\tnote{\S} & \xmark & 0.97B & 107k\tnote{\textparagraph} & - & unk. & 1162\\
    \rowcolor{pink!70} \multicolumn{9}{c}{\textbf{Fully-open models}}\\
    \rowcolor{pink!45}\multicolumn{9}{l}{\textit{AED models}}\\
    \rowcolor{pink!25}~OWSM v3.1 base~\cite{owsm31} & \cmark & \cmark & \cmark & 0.10B & 140k  & 40k & 2.3k & 151\\
    \rowcolor{pink!25}~OWSM v3.1 small~\cite{owsm31} & \cmark & \cmark & \cmark & 0.37B & 140k  & 40k & 3.2k & 151\\
    \rowcolor{pink!25}~OWSM v3.1 medium~\cite{owsm31} & \cmark & \cmark & \cmark & 1.02B & 140k  & 40k  &  24.6k & 151\\
    \rowcolor{pink!25}~OWSM v4 base (ours) & \cmark & \cmark & \cmark & 0.10B & 290k  & 30k & 1.0k\tnote{$\dagger$} & 151 \\
    \rowcolor{pink!25}~OWSM v4 small (ours) & \cmark & \cmark & \cmark & 0.37B & 290k  & 30k & 1.7k\tnote{$\dagger$} & 151 \\
    \rowcolor{pink!25}~OWSM v4 medium (ours) & \cmark & \cmark & \cmark & 1.02B & 290k  & 30k  & 3.8k\tnote{$\dagger$} & 151 \\
    \hdashline
    \rowcolor{pink!45}\multicolumn{9}{l}{\textit{CTC models}}\\
    \rowcolor{pink!25}~OWSM-CTC v3.1~\cite{owsmctc} & \cmark & \cmark & \cmark & 1.01B & 140k  & 40k  & 19.2k & 151\\
    \rowcolor{pink!25}~OWSM-CTC v3.2\tnote{$\ddagger$} & \cmark & \cmark & \cmark & 1.01B & 124k  & 30k   & 28.8k & 151\\
    \rowcolor{pink!25}~OWSM-CTC v4 (ours) & \cmark & \cmark & \cmark & 1.01B & 290k  & 30k  & 4.1k\tnote{$\dagger$} & 151\\
    \bottomrule
  \end{tabular}
  \begin{tablenotes}
    \item[*] Trained on NVIDIA A100 (80GB). Not including encoder pre-training time.
    \item[\S] Not publicly released, but provides detailed statistics and links to sources.
    \item[\textparagraph] The 491k hours of unlabeled speech for pre-training are not included here.
    \item[$\dagger$] Trained on NVIDIA H100 (96GB). Previous OWSM used A100 (40GB).
    \item[$\ddagger$] Trained on v3.1 data and then fine-tuned on v3.2~\cite{owsm32-jinchuan}, a subset of v3.1.
  \end{tablenotes}
  \end{threeparttable}
  }
  \vskip -0.15in
\end{table}
\endgroup

\begingroup
\setlength{\tabcolsep}{9pt}
\begin{table}[t]
  \caption{LID accuracy (\%) on FLERUS and long-form English ASR WER (\%) on a web presentation corpus.}
  \label{tab:lid-longasr}
  \vskip -0.13in
  \centering
  \resizebox {\linewidth} {!} {
  \begin{threeparttable}
  \begin{tabular}{lcc}
    \toprule
    \textbf{Model} & \textbf{LID Acc.} $\uparrow$ & \textbf{Long-Form WER} $\downarrow$ \\
    \midrule
    \rowcolor{gray!30} \multicolumn{3}{c}{\textbf{Open-weight models}}\\
    \rowcolor{gray!10}Whisper-medium & 54.8\tnote{*} & 3.8\\
    \rowcolor{gray!10}Whisper-large-v3 & 58.9\tnote{*} & 3.4 \\
    \rowcolor{brown!40} \multicolumn{3}{c}{\textbf{Open-weight, open-data models}}\\
    \rowcolor{brown!20}MMS-lid-4017 & 93.3 & - \\
    \rowcolor{pink!70} \multicolumn{3}{c}{\textbf{Fully-open models}}\\
    \rowcolor{pink!45}\multicolumn{3}{l}{\textit{AED models}}\\
    \rowcolor{pink!25}~OWSM v3.1 base & 41.9 & 9.6\\
    \rowcolor{pink!25}~OWSM v3.1 small & 67.1 & 6.7\\
    \rowcolor{pink!25}~OWSM v3.1 medium & 75.6 & 5.7 \\
    \rowcolor{pink!25}~OWSM v4 base (ours) & 80.1 & 5.5\\
    \rowcolor{pink!25}~OWSM v4 small (ours) & 90.0 & 4.6\\
    \rowcolor{pink!25}~OWSM v4 medium (ours) & \textbf{95.6} & 4.3\\
    \rowcolor{pink!25}~~ + beam size 5 & - & 3.6 \\
    \hdashline
    \rowcolor{pink!45}\multicolumn{3}{l}{\textit{CTC models}}\\
    \rowcolor{pink!25}~OWSM-CTC v3.1 & 87.6 & 5.2\\
    \rowcolor{pink!25}~OWSM-CTC v3.2 & 91.1 & 4.8\\
    \rowcolor{pink!25}~OWSM-CTC v4 (ours) & 93.6 & \textbf{3.3}\\
    \bottomrule
  \end{tabular}
  \begin{tablenotes}
    \item[*] Whisper supports only a subset of languages in FLEURS.
  \end{tablenotes}
  \end{threeparttable}
  }
  \vskip -0.1in
\end{table}
\endgroup

\subsection{OWSM v4 series}
\vskip -0.05in

To further assess the quality of our cleaned YODAS data, we train a new series of OWSM v4 models using this curated data alongside the previous OWSM v3.2 data~\cite{owsm32-jinchuan}.
This series includes three AED-based models ranging from 100M to 1B parameters, as well as a CTC-based model with 1B parameters.
\autoref{tab:configs} summarizes the model and training configurations.
Our v4 models employ the same configurations as the previous v3.1~\cite{owsm31, owsmctc}, except that the number of Mel filterbanks is increased from 80 to 128, following Whisper-large-v3.
The speech features are subsampled by eight times, resulting in a time shift of \SI{80}{ms}.
The speech encoder is E-Branchformer~\cite{ebf}, and the decoder, if exists, is Transformer~\cite{transformer}. 
We implement models in ESPnet~\cite{espnet} based on PyTorch~\cite{pytorch}. FlashAttention-2~\cite{flashattention2} is used for better efficiency.
We use the AdamW optimizer~\cite{adamw} with a batch size of 320. We train all models for 700k steps, i.e., around three epochs.

\begingroup
\setlength{\tabcolsep}{1.5pt}
\begin{table}[t]
  \renewcommand{\arraystretch}{1}
  \caption{Multilingual ASR WERs (\%) on MLS. The inference speed is based on the total decoding time on an NVIDIA H100.}
  \label{tab:mls}
  \vskip -0.13in
  \centering
  \resizebox {\linewidth} {!} {
  \begin{tabular}{lcccccccccc}
    \toprule
    \textbf{Model} & \texttt{eng} & \texttt{spa} & \texttt{fra} & \texttt{deu} & \texttt{nld} & \texttt{ita} & \texttt{por} & \texttt{pol} & {\footnotesize\textbf{Ave. $\downarrow$}} & {\footnotesize\textbf{Speed $\uparrow$}}\\
    \midrule
    \rowcolor{gray!30} \multicolumn{11}{c}{\textbf{Open-weight models}}\\
    \rowcolor{gray!10}Whisper-base & 13.4 & 14.5 & 25.2 & 19.9 & 30.9 & 32.9 & 23.5 & 25.2 & 23.2 & 3.9$\times$\\
    \rowcolor{gray!10}Whisper-small & 9.1 & 9.1 & 13.6 & 11.5 & 18.2 & 21.3 & 13.8 & 12.5 & 13.6 & 2.3$\times$\\
    \rowcolor{gray!10}Whisper-medium & 10.2 & 6.1 & 9.7 & 8.1 & 12.2 & 15.6 & 8.9 & 6.8 & 9.7 & 1.2$\times$\\
    \rowcolor{gray!10}Whisper-large-v3 & 5.1 & 4.1 & 4.8 & 5.6 & 10.2 & 9.2 & 7.4 & 4.4 & 6.4 & 1.0$\times$\\
    \rowcolor{brown!40} \multicolumn{11}{c}{\textbf{Open-weight, open-data models}}\\
    \rowcolor{brown!20} MMS-fl102 & 23.6 & 14.9 & 22.4 & 14.7 & 16.4 & 18.9 & 17.1 & 12.7 & 17.6 & 20.9$\times$\\
    \rowcolor{brown!20} MMS-all & 10.7 & 5.8 & 8.8 & 8.8 & 12.8 & 11.0 & 16.2 & 10.5 & 10.6 & 21.4$\times$\\
    \rowcolor{pink!70} \multicolumn{11}{c}{\textbf{Fully-open models}}\\
    \rowcolor{pink!45}\multicolumn{11}{l}{\textit{OWSM-AED models}}\\
    \rowcolor{pink!25}~v3.1 base & 12.0 & 18.5 & 24.2 & 18.7 & 28.6 & 33.7 & 44.9 & 49.7 & 28.8 & 3.0$\times$\\
    \rowcolor{pink!25}~v3.1 small & 8.1 & 10.8 & 14.1 & 12.4 & 19.7 & 21.8 & 26.7 & 28.5 & 17.8 & 2.2$\times$\\
    \rowcolor{pink!25}~v3.1 medium & 7.1 & 9.0 & 12.1 & 10.8 & 18.1 & 20.2 & 21.6 & 25.2 & 15.5 & 1.2$\times$\\
    \rowcolor{pink!25}~v4 base (ours) & 11.6 & 11.6 & 17.6 & 15.9 & 23.1 & 23.3 & 18.9 & 31.5 & 19.2 & 3.0$\times$\\
    \rowcolor{pink!25}~v4 small (ours) & 7.6 & 7.1 & 10.2 & 10.3 & 15.7 & 15.7 & 11.5 & 16.0 & 11.8 & 2.2$\times$\\
    \rowcolor{pink!25}~v4 medium (ours) & 6.4 & 5.7 & 7.8 & 8.2 & 13.4 & 13.1 & 9.0 & 11.5 & 9.4 & 1.1$\times$\\
    \rowcolor{pink!25}~~+ beam size 5 & 5.9 & 5.5 & 7.3 & 7.9 & 12.9 & 12.8 & 8.5 & 11.0 & 9.0 & 0.2$\times$\\
    \hdashline
    \rowcolor{pink!45}\multicolumn{11}{l}{\textit{OWSM-CTC models}}\\
    \rowcolor{pink!25}~v3.1 & 7.3 & 10.3 & 12.9 & 11.9 & 20.4 & 22.1 & 23.5 & 31.6 & 17.5 & 26.3$\times$\\
    \rowcolor{pink!25}~v3.2 & 7.0 & 9.7 & 11.3 & 11.4 & 17.6 & 20.0 & 20.5 & 24.5 & 15.3 & 23.7$\times$\\
    \rowcolor{pink!25}~v4 (ours) & 6.4 & 5.8 & 7.8 & 9.5 & 15.1 & 15.5 & 10.3 & 15.1 & 10.7 & 25.1$\times$\\
    \bottomrule
  \end{tabular}
  }
  \vskip -0.19in
\end{table}
\endgroup

\begingroup
\setlength{\tabcolsep}{2pt}
\begin{table*}[t]
  \caption{English ASR WERs (\%) on the Hugging Face Open ASR Leaderboard. The Inverse Real Time Factor (RTFx) is measured using an NVIDIA H100 GPU (96GB). \underline{Underlined}: Our v4 model outperforms previous OWSM.
  }
  \label{tab:hf-asr-leaderboard}
  \vskip -0.12in
  \centering
  \resizebox {\linewidth} {!} {
  \begin{tabular}{lcccccccccccc}
    \toprule
    \textbf{Model} & \textbf{Arch.} & \textbf{Size} & \textbf{Ave. WER} $\downarrow$ & \textbf{RTFx} $\uparrow$ & \textbf{AMI} & \textbf{Earnings22} & \textbf{Gigaspeech} & \textbf{LS-Clean} & \textbf{LS-Other} & \textbf{SPGISpeech} & \textbf{Web-Presentation} & \textbf{Voxpopuli}\\
    \midrule
    \rowcolor{gray!30} \multicolumn{13}{c}{\textbf{Open-weight models}}\\
    \rowcolor{gray!10}Whisper-medium-en & AED & 0.8B & 8.06 & 289.64 & 16.66 & 12.42 & 11.11 & 2.89 & 5.85 & 3.36 & 4.13 & 8.04\\
    \rowcolor{gray!10}Whisper-large-v3 & AED & 1.6B & 7.47 & 235.61 & 16.00 & \textbf{11.39} & \textbf{10.10} & 2.01 & 3.92 & 2.95 & 3.84 & 9.51\\
    \rowcolor{gray!10}Canary & AED & 1.0B & \textbf{6.48} & 287.62 & 13.66 & 12.19 & 10.12 & \textbf{1.47} & \textbf{2.96} & \textbf{2.06} & 3.59 & \textbf{5.81}\\
    \rowcolor{gray!10}Parakeet-CTC & CTC & 1.1B & 7.40 & \textbf{3007.88} & 15.66 & 13.77 & 10.28 & 1.86 & 3.50 & 4.02 & \textbf{3.54} & 6.55\\
    \rowcolor{brown!40} \multicolumn{13}{c}{\textbf{Open-weight, open-data models}}\\
    \rowcolor{brown!20}MMS-fl102 & CTC & 1.0B & 39.90 & 1066.12 & 86.80 & 51.74 & 42.44 & 22.13 & 28.76 & 26.21 & 32.35 & 28.80\\
    \rowcolor{brown!20}MMS-all & CTC & 1.0B & 22.65 & 1055.91 & 42.02 & 31.19 & 26.44 & 12.64 & 15.98 & 16.95 & 17.49 & 18.50\\
    \rowcolor{pink!70}\multicolumn{13}{c}{\textbf{Fully-open models}}\\
    \rowcolor{pink!25}OWSM-CTC v3.1 & CTC & 1.0B & 8.12 & 853.44 & 15.66 & 13.73 & 11.89 & 2.36 & 5.12 & 2.87 & 4.97 & 8.36\\
    \rowcolor{pink!25}OWSM-CTC v3.2 & CTC & 1.0B & 8.24 & 841.12 & 16.71 & 13.50 & 11.78 & 2.61 & 5.32 & 2.73 & 5.35 & 7.95\\
    \rowcolor{pink!25}OWSM-CTC v4 (ours) & CTC & 1.0B & \underline{7.44} & 791.18 & \underline{\textbf{13.09}} & 13.89 & \underline{10.83} & 2.56 & \underline{4.86} & \underline{2.56} & \underline{4.40} & \underline{7.34}\\
    \bottomrule
  \end{tabular}
  }
  \vskip -0.05in
\end{table*}
\endgroup

\begingroup
\setlength{\tabcolsep}{3pt}
\begin{table*}[t]
  \caption{BLEU scores (\%) for ST on CoVoST-2~\cite{covost2}. We do not add any new ST data; OWSM-CTC v4 uses the same ST data as v3.2.}
  \label{tab:st-all}
  \vskip -0.13in
  \centering
  \resizebox {\linewidth} {!} {
  \begin{tabular}{lccccccccccccccccccccc}
    \toprule
    \multirow{2}{*}{\textbf{Model}} & \multicolumn{5}{c}{\textbf{X-En Translation}} & \multicolumn{16}{c}{\textbf{En-X Translation}} \\
    \cmidrule(lr){2-6}
    \cmidrule(lr){7-22}
    & \texttt{de} & \texttt{es} & \texttt{fr} & \texttt{ca} & \textbf{Average} & \texttt{de} & \texttt{ca} & \texttt{zh} & \texttt{fa} & \texttt{et} & \texttt{mn} & \texttt{tr} & \texttt{ar} & \texttt{sv} & \texttt{lv} & \texttt{sl} & \texttt{ta} & \texttt{ja} & \texttt{id} & \texttt{cy} & \textbf{Average} \\
    \midrule
    OWSM v3.1 & 16.7 & 22.3 & 22.8 & 18.8 & 20.2 & 26.3 & 20.4 & 29.7 & 10.2 & 9.6 & 5.8 & 7.8 & 7.2 & 20.8 & 8.4 & 11.0 & 0.1 & \textbf{21.1} & 17.2 & 16.3 & 14.1 \\
    OWSM-CTC v3.1 & 20.7 & 27.9 & 27.5 & 24.2 & 25.1 & 26.7 & 24.0 & 32.9 & 9.9 & 11.4 & 6.2 & 7.9 & 8.3 & 24.5 & 10.0 & 14.2 & 0.1 & 20.4 & 22.6 & 20.6 & 16.0 \\
    OWSM-CTC v3.2 & 21.1 & 28.6 & 27.6 & 24.2 & 25.4 & 27.8 & \textbf{25.2} & 33.4 & \textbf{11.0} & \textbf{12.0} & \textbf{6.7} & 8.9 & \textbf{9.7} & \textbf{26.0} & \textbf{11.3} & \textbf{15.1} & 0.1 & 20.9 & 24.0 & \textbf{21.5} & \textbf{16.9} \\
    OWSM-CTC v4 (ours) & \textbf{22.1} & \textbf{31.8} & \textbf{29.5} & \textbf{25.7} & \textbf{27.3} & \textbf{28.5} & 24.9 & \textbf{35.3} & 10.6 & 11.6 & 6.3 & \textbf{9.2} & 9.2 & 24.9 & 10.2 & 14.3 & 0.0 & 20.8 & \textbf{24.6} & 19.8 & 16.7 \\
    \bottomrule
  \end{tabular}
  }
  \vskip -0.2in
\end{table*}
\endgroup

\section{Experimental Results}
\vskip -0.05in

We evaluate our OWSM v4 models on multilingual ASR, LID, and ST benchmarks using greedy decoding unless otherwise specified. While we include results from models developed by well-resourced industry entities such as OpenAI's Whisper and Meta's MMS, our primary comparisons are against baselines from academic institutions, given our constrained resources.

\subsection{Results of language identification}
\vskip -0.07in

\autoref{tab:lid-longasr} presents the LID results on FLEURS~\cite{FLEURS}, where our OWSM v4 series outperforms earlier versions. Compared to industrial-scale models, OWSM v4 medium and OWSM-CTC v4 both achieve higher accuracies than Whisper and MMS-lid, with OWSM v4 medium reaching the highest accuracy of 95.6\%.
These results indicate that our cleaned YODAS data contains high-quality language labels, attributed to the LID filtering stage (see Section~\ref{subsubsec:lid-filtering}).

\subsection{Results of multilingual speech recognition}
\label{subsec:results-multilingual}
\vskip -0.05in

\autoref{tab:mls} presents ASR results on MLS~\cite{pratap2020mls}. Again, our OWSM v4 series achieves much lower WERs than previous OWSM of the same size across all eight languages, highlighting the benefit of data scaling and cleaning.
Compared to leading industrial models, OWSM v4 medium achieves a lower average WER than Whisper-medium (9.4\% vs. 9.7\%) with a similar inference speed.
OWSM-CTC v4 achieves a much lower WER than MMS-fl102 (10.7\% vs. 17.6\%) and a similar WER to MMS-all (10.7\% vs. 10.6\%), while being 20\% faster.

We also evaluate OWSM-CTC on FLEURS~\cite{FLEURS}.\footnote{The raw transcriptions in FLEURS include words in parentheses, some of which are spoken while others are not. As there is no straightforward rule to exclude non-spoken words, we use the ``transcription'' field from the Hugging Face dataset as the groundtruth. This may result in higher error rates for certain languages, such as Chinese.}
OWSM-CTC v4 outperforms v3.1 in all 102 languages and surpasses v3.2 in 100 languages.
\autoref{tab:fleurs} shows 21 languages where OWSM-CTC v4 has error rates below 20\%. 
Among them, OWSM-CTC v4 outperforms MMS-all in 13 languages.
These findings further validate the effectiveness of our approach.

\subsection{Results of English speech recognition}
\vskip -0.05in

\autoref{tab:hf-asr-leaderboard} presents English ASR WERs on the Hugging Face Open ASR leaderboard~\cite{open-asr-leaderboard}.\footnote{ESPnet does not support batched beam search, leading to very slow inference for AED models. Hence, we only decode CTC-based OWSM.}
Our OWSM-CTC v4 outperforms previous OWSM-CTC on 6 of 8 test sets. The average WER is improved from 8.12\% to 7.44\%.
Our model also significantly surpasses MMS-fl102 and MMS-all with a similar size.
Compared to leading industrial models trained on proprietary data, our model outperforms Whisper-medium while achieving performance on par with Whisper-large-v3 and Parakeet-CTC.
Regarding inference speed, our OWSM-CTC v4 is several times faster than AED models such as Whisper and Canary, consistent with the findings in~\cite{owsmctc}.\footnote{Unlike NeMo or Hugging Face's transformers, ESPnet lacks lower-level optimization for inference. Nevertheless, our model still achieves competitive inference speed.}

\autoref{tab:lid-longasr} shows long-form English ASR results, where our OWSM v4 models significantly outperform previous OWSM v3.1 and v3.2 of the same size and category (AED or CTC).
Notably, OWSM v4 base (100M) already surpasses OWSM v3.1 medium (1B).
Compared to frontier industrial models, OWSM-CTC v4 achieves the lowest long-form WER of 3.3\%, slightly outperforming Whisper-large-v3, which has 50\% more parameters and is trained on 15 times more data.
These findings highlight the quality of our curated English data from YODAS and demonstrate the benefit of data scaling.

\subsection{Results of speech translation}
\vskip -0.07in

We do not add any new ST data, using exactly the same ST data as v3.2.
Here, our goal is to show that our v4 model maintains similar ST performance.
Following \cite{owsmctc}, we evaluate ST performance on CoVoST-2 X-En and En-X~\cite{covost2}. 
As shown in \autoref{tab:st-all}, OWSM-CTC v4 achieves higher BLEU scores than previous OWSM in the four X-En test sets and comparable scores to v3.2 in En-X test sets, verifying that using additional ASR data from YODAS does not negatively impact ST performance.

\section{Conclusion}
\vskip -0.07in

We improve fully open speech-to-text foundation models via data scaling and cleaning using academic-scale resources.
We reveal that large-scale web-crawled data contains incorrect language labels and audio-text misalignments. 
To mitigate these issues, we develop a scalable data-cleaning pipeline using public models and toolkits. 
Applying it to the raw YODAS ASR dataset, we create a higher-quality subset with 166k hours of speech in 75 languages. 
Furthermore, we train a new series of OWSM v4 models using this curated dataset alongside existing OWSM data. 
Extensive evaluations show that our models consistently and significantly outperform previous OWSM models on multilingual benchmarks. Our models even match or surpass leading industrial models such as Whisper and MMS on multiple benchmarks. To advance open academic research, we will publicly release our data-cleaning scripts, the curated YODAS dataset, training code, pre-trained models, and training logs.

\vspace{-0.05in}
\section{Acknowledgements}
\vskip -0.05in

We use PSC Bridges2 and NCSA Delta via ACCESS CIS210014, by National Science Foundation grants \#2138259, \#2138286, \#2138307, \#2137603, and \#2138296.

\section{References}
\printbibliography

\end{document}